\documentclass[conference]{IEEEtran}
\IEEEoverridecommandlockouts
\usepackage{cite}
\usepackage{amsmath,amssymb,amsfonts}
\usepackage{algorithmic}
\usepackage{graphicx}
\usepackage{textcomp}
\usepackage{xcolor}
\usepackage{float}
\usepackage{url} 
\usepackage{multirow}
\usepackage{pdflscape}
\usepackage{adjustbox}
\usepackage{tabularx}

\def\BibTeX{{\rm B\kern-.05em{\sc i\kern-.025em b}\kern-.08em
    T\kern-.1667em\lower.7ex\hbox{E}\kern-.125emX}}
\begin{document}
\title{
Indian Legal Text Summarization: A Text Normalisation-based Approach
}
\author{\IEEEauthorblockN{Satyajit Ghosh}
\IEEEauthorblockA{\textit{Department of CSE} \\
\textit{Adamas University}\\
Kolkata, India \\
satyajit.ghosh@stu.adamasuniversity.ac.in}
\and
\IEEEauthorblockN{Mousumi Dutta}
\IEEEauthorblockA{\textit{Department of CSE} \\
\textit{Adamas University}\\
Kolkata, India \\
mousumidutta@stu.adamasuniversity.ac.in}
\and
\IEEEauthorblockN{Tanaya Das}
\IEEEauthorblockA{\textit{Department of CSE} \\
\textit{Adamas University}\\
Kolkata, India \\
tanayadas.das23@gmail.com}
}
\IEEEoverridecommandlockouts

\IEEEpubid{
\resizebox{\columnwidth}{!}{%
\begin{tabular}{l}
\begin{tabular}[c]{@{}l@{}}© 2022 IEEE. Personal use of this material is permitted. Permission from IEEE must be \\ obtained for all other uses, in any current or future media, including \\ reprinting/republishing this material for advertising or promotional purposes, creating new \\ collective works, for resale or redistribution to servers or lists, or reuse of any copyrighted \\ component of this work in other works.\\
\\ \textbf{Accepted at 2022 IEEE 19th India Council International Conference (INDICON)}
\end {tabular}
\end{tabular}%
}
\hspace{\columnsep}\makebox[\columnwidth]{ }}
\maketitle
\IEEEpubidadjcol
\begin{abstract}
In the Indian court system, pending cases have long been a problem. There are more than 4 crore cases outstanding. Manually summarising hundreds of documents is a time-consuming and tedious task for legal stakeholders. Many state-of-the-art models for text summarization have emerged as machine learning has progressed. Domain-independent models don't do well with legal texts, and fine-tuning those models for the Indian Legal System is problematic due to a lack of publicly available datasets. To improve the performance of domain-independent models, the authors have proposed a methodology for normalising legal texts in the Indian context. The authors experimented with two state-of-the-art domain-independent models for legal text summarization, namely BART and PEGASUS. BART and PEGASUS are put through their paces in terms of extractive and abstractive summarization to understand the effectiveness of the text normalisation approach. Summarised texts are evaluated by domain experts on multiple parameters and using ROUGE metrics. It shows the proposed text normalisation approach is effective in legal texts with domain-independent models.

\end{abstract}


\section{Introduction}
Text summarization is the process of constructing a concise, cohesive, and fluent summary of a lengthy text document \cite{Popescu2020}. It gives us a brief context of the story. Statutes (established laws) and Precedents (prior cases) are the two primary sources of law for countries that follow the Common Law System like India \cite{Hiware2019}. Hence, there are hundreds of prior cases that lawyers must go through. Legal documents can be lengthy. The abbreviations and terminology used in Legal documents are different from the standard language.  Manual drafting of case summaries is a process that takes a lot of time. Automatic summarization of texts is possible because of the advancement of Artificial Intelligence (AI) and Machine Learning (ML). Thousands of hours of man-labour can be reduced with the help of State-of-the-Art Machine Learning models. It is also useful for beginners and ordinary citizens to understand a judgement. More than 4.70 crore cases are pending in various courts in India \cite{PTI2022}. The use of automatic summarization of texts will also significantly reduce the number. Several legal text summarization techniques and tools have been reported in the past based on UK \cite{Grover2003}, Canadian \cite{Farzindar2004_let} and Australian \cite{Polsley2016_1} court judgements. As each country has its structure and abbreviations in their Legal documents it is not suitable to use those tools and techniques for other countries. To train a model for domain-specific text summarization, we need a lot of data. As there is no publicly available dataset for summarization of Indian legal documents present, so we have proposed a different methodology to summarize them. Text summarization techniques are classified into two categories. An abstractive summarization recognises the language in the text and adds novel words to the summary if necessary \cite{Zhu2021,Moratanch2016} and in extractive summarization, a summary is formed from a subset of sentences \cite{Moratanch2017}. In this paper, The Indian legal texts have been normalised as a general text based on our proposed methodology. Next, two domain-independent models have been used for summarization. The authors have tried to conduct extractive summarization using BART \cite{Lewis2019} and abstractive summarization with PEGASUS \cite{Zhang2019}.
Section II presents the related work behind the proposed methodology. Section III discusses a detailed explanation of the proposed methodology. Section IV presents the evaluation of the work compared to the traditional one. Finally, Section V concludes the paper along with its future applications.
\section{Related Work}

Legal case judgements are usually lengthy and complicated due to the use of many domain-specific abbreviations \cite{Hiware2019}. There is a vast amount of research conducted for Legal text summarization in different countries like the US, Canada, Australia, and the UK. Most of the research works tried to train deep-learning models using a supervised or semi-supervised approach for legal text summarization. J. W. Yingjie and Ma in \cite{Ju2013} selected three sentences from each topic that are having best representation of the topic. The proposed summarization method blends term description with sentence description for each topic using LSA (Latent Semantic Analysis). B. Samei et al. in \cite{Samei2014} introduced a model for multi-document summarization using graph-based and information-theoretic concepts. A. Farzindar and G. Lapalme in \cite{Farzindar2004} proposed legal documents summarization based on the exploration of the documents’ architecture and thematic structures. A. Joshi et al. in \cite{Joshi2019} describes a summary based on the “three-sentence selection” metrics. These are “content relevance”, “sentence novelty”, and “sentence position relevance”. The sentence content relevance is measured using a deep auto-encoder network. S. Polsley et al. in \cite{Polsley2016} introduced a tool which takes advantage of standard summary methods based on word frequency and generates automated text summarization of the legal documents. The tool is evaluated using Recall-Oriented Understudy for Gisting Evaluation (ROUGE) and human scoring. Vijayasanthi et al. in  \cite{Ka2013} proposed a hybrid system for automatic text summarization of legal documents. Key phrase matching and case-based techniques are involved in the hybrid system. M. Saravanan and B. Ravindran in  \cite{Saravanan2010} proposed a system for labelling sentences with their rhetorical roles. The application of probabilistic models for extraction of the key sentences is described by them. N. Bansal et al. in \cite{Bansal2019} introduced a Fuzzy Analytical Hierarchical Process (FAHP) based feature weighting scheme for producing summaries of legal judgements. They found it to be more promising than other traditional approaches.
State-of-the-Art domain-independent models are not tried in the past research for legal text summarization. The objective is to find out the effectiveness of State-of-the-Art domain-independent models for domain-specific tasks like legal case summarization. Its findings will help in legal and other sectors where the models can’t be properly trained or fine-tuned due to a lack of data. In the next section, the authors discuss the detailed methodology for text summarization using State-of-the-Art domain-independent models.

\section{Methodology}
The public records of the Indian judiciary are disorganised and noisy \cite{Parikh2021}. There is no publicly available dataset of Indian legal documents summarization. So, the authors proposed a novel methodology without the need for a dataset.
%% Figure 1

\begin{figure}[H]
    \centering
    \includegraphics[width=9cm]{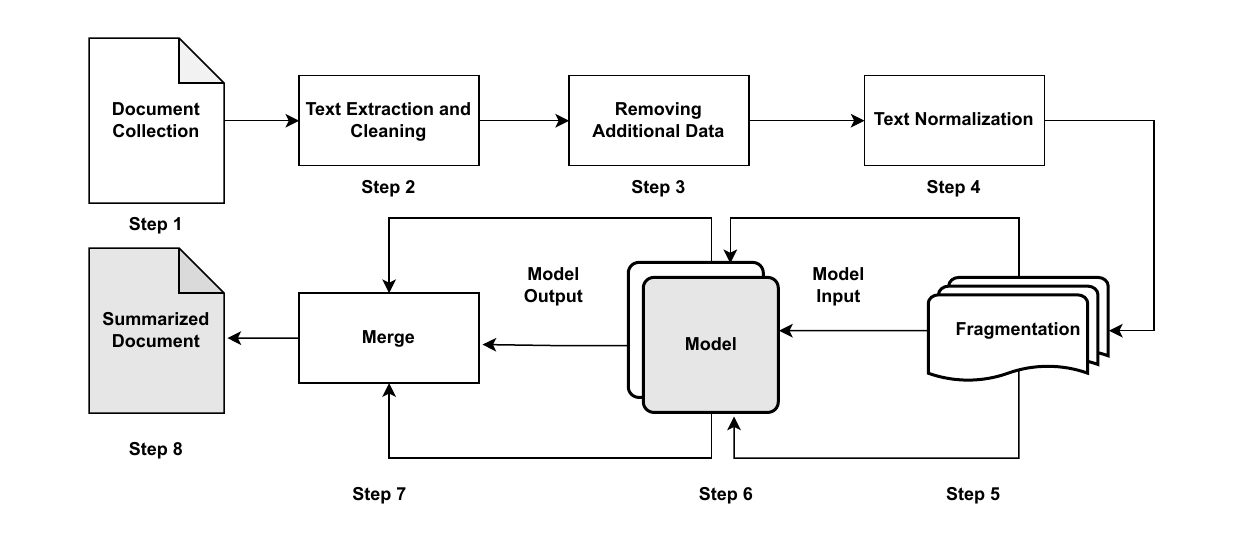}
    \caption{Block diagram of the proposed methodology}
\end{figure}

Fig.1. depicts our proposed methodology. In step 1 we collected the Indian Legal documents from sources such as SCI, IndianKanoon, Manupatra and ILDC \cite{Malik2021}. In step 2, we have extracted the texts using Optical Character Recognition (OCR) from the documents. Next, the noise is removed from the extracted text using basic clean up techniques e.g., White-space removal, Spelling correction. After that, in step 3 the additional information at the beginning before the actual judgement is removed.  Then the Legal texts are normalised in step 4. We have constructed a dictionary with Legal abbreviations and their full form. A separate dictionary is created for articles, sections, and their summary. We have replaced the Legal abbreviations with their full form and appended the article and section mentions with its summary in the documents. Table I displays the raw legal texts received from the documents and after step 4 their normalised versions.

%% Table 1

\begin{table}[]
\centering
\caption{Raw and Normalised Texts}
\begin{tabularx}{\linewidth}{|l|X|}
\hline
\multicolumn{2}{|l|}{Example 1} \\ \hline
\multicolumn{1}{|l|}{Raw Legal Text} &
  “. . . they approached the High Court of Kerala by way of W.P. (C) No. 2329 of 2014.” \\ \hline
\multicolumn{1}{|l|}{Normalised Text} &
  “. . . they approached the High Court of Kerala by way of \textbf{Writ Petition under 226 and 227 of the Constitution} No. 2329 of 2014.” \\ \hline
\multicolumn{2}{|l|}{Example 2} \\ \hline
\multicolumn{1}{|l|}{Raw Legal Text} &
  “. . . petition filed by the appellant under Section 13 of Hindu Marriage Act,1955 on grounds of desertion.” \\ \hline
\multicolumn{1}{|l|}{Normalised Text} &
  “. . . petition filed by the appellant under Section 13 \textbf{(for divorce)} of Hindu Marriage Act,1955 on grounds of desertion.” \\ \hline
\end{tabularx}
\end{table}

After normalisation, the text is divided into small fragments in step 5, and then those fragments are given to the model in step 6. If the entire document is supplied at once, the model is unable to understand and extract key points from it. Thus the inputs are provided in small fragments to these models. Those outputs later merged for the actual summarised document.We have used two models. The first one is BART. It is used for extractive summarization. BART is described as a denoising autoencoder which is implemented as a sequence-to-sequence model and a bi-directional encoder \cite{Lewis2019}.
BART can be used in many downstream applications. K. M. Hermann et al. in \cite{Hermann2015} proposed a new methodology to fine-tune BART for Comprehension, Translation and Natural Language Generation with very less amount of training data. The model is built for summarization in English texts based on that paper.
The second model is PEGASUS. It is a pre-trained model with extracted Gap-sentences for abstractive summarization. Good abstractive summarization performance can be achieved across a broad domain with very little supervision by fine-tuning PEGASUS \cite{Zhang2019}.
Next, in step 7 the model outputs for all the small fragments are merged and in step 8 finally, we get the summarised document.
The model outputs at the next stage are evaluated by different Legal experts on multiple parameters.

\section{Evaluation}
The traditional way of evaluating summaries is to use the ROUGE scores or expert evaluation. We have calculated  ROUGE-1, ROUGE-2, ROUGE-3 and ROUGE-L scores . To calculate the scores we have compared model summaries generated from raw texts and normalised texts with the summaries provided by Legal experts.

\begin{table}[H]
\caption{Average ROUGE scores after comparing Normalised text}
\centering
\begin{tabular}{|l|l|l|l|}
\hline
\textbf{Type} & \textbf{Precision} & \textbf{Recall} & \textbf{F-Score} \\
\hline
ROUGE-1       & 0.46               & 0.55            & 0.48             \\
ROUGE-2       & 0.28               & 0.36            & 0.31             \\
ROUGE-3       & 0.21               & 0.27            & 0.24             \\
ROUGE-L       & 0.35               & 0.45            & 0.39            \\
\hline
\end{tabular}
\end{table}
\begin{table}[H]
\caption{Average ROUGE scores after comparing Raw text}
\centering
\begin{tabular}{|l|l|l|l|}
\hline
\textbf{Type} & \textbf{Precision} & \textbf{Recall} & \textbf{F-Score} \\
\hline
ROUGE-1       & 0.45               & 0.4             & 0.42             \\
ROUGE-2       & 0.26               & 0.24            & 0.25             \\
ROUGE-3       & 0.18               & 0.17            & 0.18             \\
ROUGE-L       & 0.33               & 0.3             & 0.31            \\
\hline
\end{tabular}
\end{table}

Table II and Table III shows the performances of summarizing the samples
in terms of ROUGE-1, ROUGE-2, ROUGE-3 and ROUGE-L Precision, Recall and F-scores. Comparing these two tables we can observe that average scores on all the three parameters are increased in normalised texts. The comparison is done using BART outputs as PEGASUS is evaluated poorly by legal experts which are discussed below. Perfect scores for extractive summarization are both theoretically and computationally very difficult to achieve using ROUGE \cite{Schluter2017}. Thus we have also evaluated our results by experts.
Random samples have been taken from our data sources and as per our proposed methodology, those samples are processed. We have sent those outputs to Legal experts for evaluation. We asked them to evaluate the summaries based on three parameters on a scale of 1 to 10. The parameters are conciseness, accuracy, and detail preservation. PEGASUS model is only able to summarize the contents with more than 7-point conciseness and accuracy in 20\% samples. In some samples, it gives completely out of context summaries. In the case of BART on over 60\% of the samples, it received more than a 7-point score in all the 3 parameters from the Legal experts. Fig.2. depicts average scores given by Legal experts on different parameters in our samples.
%% Figure 2

\begin{figure}[h!]
    \centering
    \includegraphics[width=8cm]{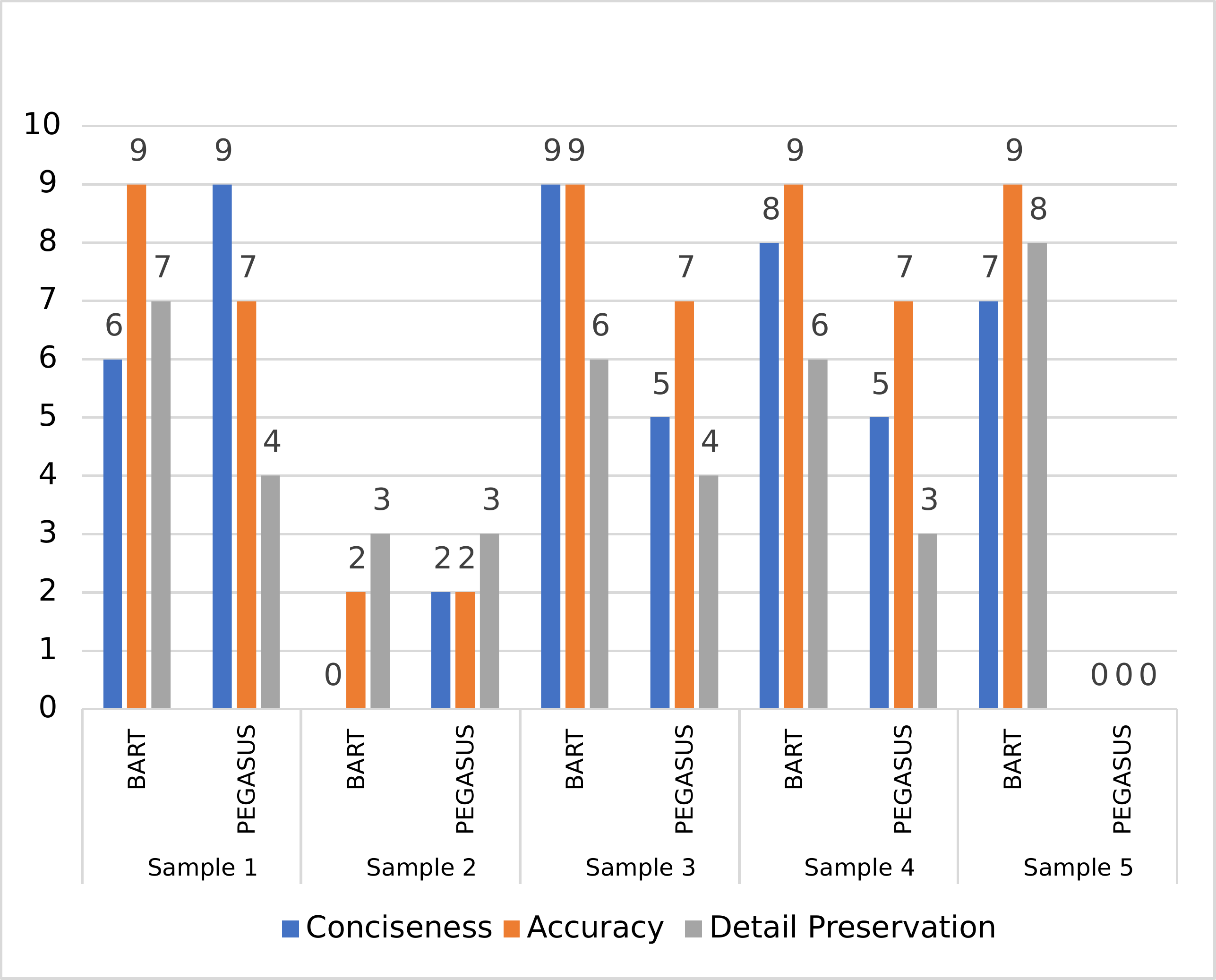}
    \caption{Evaluation by Legal Experts}
\end{figure}
 Both normalised text and raw Legal texts are provided to the BART model and the summary outputs are then shown to naive persons. The summaries from the normalised texts are evaluated as more conscious and easy to understand by them. Table IV compares the raw and normalised legal text summaries. 

%% Table 2
\begin{table}[]
\centering
\caption{Summary from Raw and Normalised Legal Texts}
\begin{tabularx}{\linewidth}{|l|X|}
\hline
\multicolumn{2}{|l|}{Example 1} \\ \hline
\multicolumn{1}{|l|}{Raw Texts Summary} &
  “The road was constructed but no compensation was paid. The Panchayat denied of having given any assurance regarding adequate compensation to be paid to the appellants.” \\ \hline
\multicolumn{1}{|l|}{Normalized Texts Summary} &
  “The road was constructed but no compensation was paid. The appellants made various representations starting from the time, construction was going on and even after the construction work was completed. When no heed was paid to their request, they approached the High Court of Kerala.” \\ \hline
\multicolumn{2}{|l|}{Example 2} \\ \hline
\multicolumn{1}{|l|}{Raw Texts Summary} &
  “The Family Court ruled that the marriage between the parties is irretrievably broken down. The appeal is also pending before this Court for the last 12 years (@ SLP of the year 2010). The appellant is stated to have got married after the decree of divorce was granted.” \\ \hline
\multicolumn{1}{|l|}{Normalized Texts Summary} &
  “The Family Court ruled that the marriage between the parties is irretrievably broken down. The appeal is also pending before this Court for the last 12 years (@ Special Leave Petition of the year 2010). The appellant is stated to have got married after the divorce was granted.” \\ \hline
\end{tabularx}
\end{table}

As shown in Fig. 3. on average 75\% decrease in the length of the normalized text is observed by summarization using BART from our samples.
%% Figure 3

\begin{figure}[H]
    \centering
    \includegraphics[width=8cm]{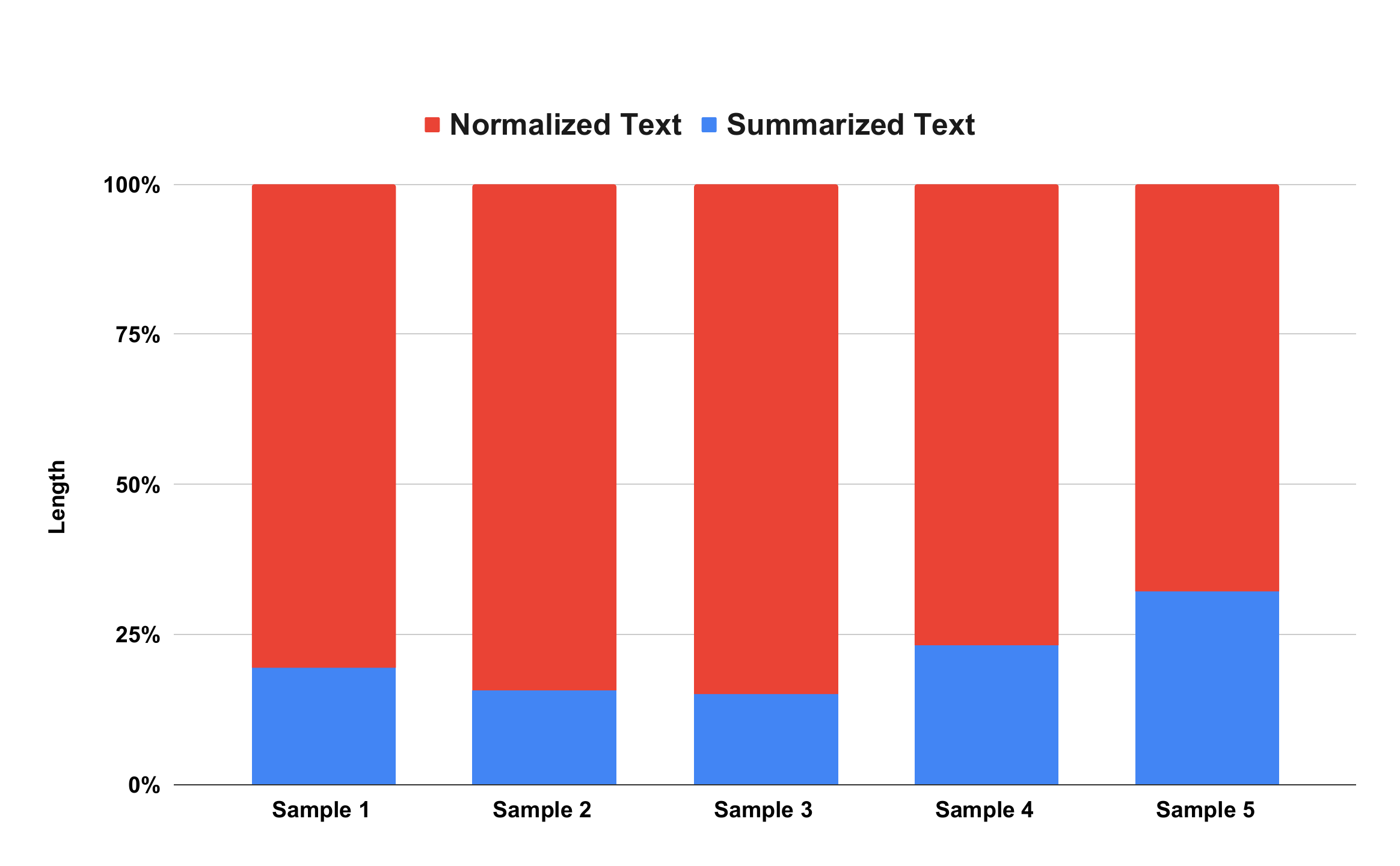}
    \caption{Summarized vs. Normalized Legal Text Length}
\end{figure}

It can significantly improve the efficacy of the legal system. Beginners and common people can also take advantage of it to understand different aspects of a legal case without any specific domain knowledge.

\section{Conclusion}
In this paper, the authors have experimented with two State-of-the-Art machine learning models for Indian Legal text summarization. BART performed well in summarizing Legal texts and decreased the length of the document up to 75\%. PEGASUS is used for abstractive summarization, but it did not work well most of the time. Thus, the normalization methodology is effective for extractive summarization but certainly not that much useful for abstractive summarization. The ROUGE metrics and expert evaluation show that even without domain-specific training State-of-the-Art machine learning models can be used for various domain-specific fields after normalizing the raw texts. All the supplementary files are available at \url{https://github.com/SATYAJIT1910/ILDS}.

\section*{Acknowledgment}
We are grateful to Prasenjit Das from the Department of Law, Adamas University and Souvik Ghosh from Amity Law School, Kolkata for evaluating our results.

%Bibliography
\bibliographystyle{ieeetr}  
\bibliography{references}

\end{document}